\def\year{2022}\relax
\documentclass[letterpaper]{article} 
\usepackage{aaai22}  
\usepackage{times}  
\usepackage{helvet}  
\usepackage{courier}  
\usepackage[hyphens]{url}  
\usepackage{graphicx} 
\usepackage{natbib}  
\usepackage{caption} 
\DeclareCaptionStyle{ruled}{labelfont=normalfont,labelsep=colon,strut=off} 
\usepackage[table,xcdraw]{xcolor}  
\usepackage{amsmath, amssymb,amsthm}
\usepackage{bbold}
\usepackage[english]{babel}
\usepackage{graphicx}
\usepackage{booktabs}
\usepackage[autostyle]{csquotes}
\usepackage[capitalise]{cleveref}
\usepackage[export]{adjustbox}
\usepackage{algpseudocode}
\usepackage{blindtext}
\usepackage{bm}
\usepackage{subfig}
\usepackage{diagbox}
\usepackage{multirow}
\usepackage[ruled,vlined]{algorithm2e}
\SetKwInput{kwInput}{Input}
\usepackage{float}

\newtheorem{proposition}{Proposition}

\usepackage{amsmath,amsfonts,bm}

\def\eqref#1{equation~\ref{#1}}

\def\1{\bm{1}}

\def\vm{{\bm{m}}}

\def\vw{{\bm{w}}}
\def\vx{{\bm{x}}}
\def\vy{{\bm{y}}}

\DeclareMathAlphabet{\mathsfit}{\encodingdefault}{\sfdefault}{m}{sl}
\SetMathAlphabet{\mathsfit}{bold}{\encodingdefault}{\sfdefault}{bx}{n}

\expandafter\def\expandafter\normalsize\expandafter{%
    \normalsize
    \setlength\abovedisplayskip{6pt}
    \setlength\belowdisplayskip{6pt}
    \setlength\abovedisplayshortskip{6pt}
    \setlength\belowdisplayshortskip{6pt}
}
\usepackage{newfloat}
\usepackage{listings}
\floatstyle{ruled}
\newfloat{listing}{tb}{lst}{}
\floatname{listing}{Listing}
\title{GROWN: GRow Only When Necessary for Continual Learning}
\author{Li Yang, Sen Lin, Junshan Zhang, Deliang Fan}
\affiliations{
School of Electrical, Computer and Energy Engineering, Arizona State University, Tempe, AZ\\
\tt\small \{lyang166, slin70, Junshan.Zhang, dfan\}@asu.edu}
\usepackage{bibentry}

\begin{document}

\maketitle

\begin{abstract}
\textit{Catastrophic forgetting} is a notorious  issue in deep learning, referring to the fact that  Deep Neural Networks (DNN) could forget the knowledge about earlier tasks when learning new tasks.
To address this issue, continual learning has been developed to  learn new tasks sequentially and perform knowledge transfer from the old tasks to the new ones without forgetting. While recent structure-based learning methods show the capability of alleviating the forgetting problem, these methods start from a redundant full-size network and require a complex learning process to gradually grow-and-prune or search the network structure for each task, which is inefficient. To address this problem and enable efficient network expansion for new tasks, we first develop a learnable sparse growth method eliminating the additional pruning/searching step in previous structure-based methods. Building on this learnable sparse growth method, we then propose \textit{GROWN}, a novel end-to-end continual learning framework to \textit{dynamically grow the model only when necessary}. Different from all previous structure-based methods,  GROWN starts from a small seed network, instead of a full-sized one. We validate GROWN on multiple dataset against
state-of-the-art methods, which shows superiority performance in both accuracy and model size. For example, we achieve 1.0$\%$ accuracy gain in average compared to the current SOTA results on CIFAR-100 Superclass 20 tasks setting. 
\end{abstract}

\section{Introduction}
It is well-known that human can learn new tasks without forgetting old ones. However, Deep Neural Networks (DNN) may forget  the ``old'' knowledge when it is trained to learn a single task. For example, given a backbone DNN model, conventional model fine-tuning for new tasks could easily result in the forgetting of old knowledge, and thus degrade the learning performance of the model on earlier tasks. Such a phenomenon is known as \textit{catastrophic forgetting}. To address this problem and enable the continuous learning as human beings do, continual learning (CL)~\cite{kirkpatrick2017overcoming}, a.k.a, lifelong learning, has recently attracted much attention. It aims to build a model that is incrementally updated over a sequence of tasks, performing knowledge transfer from the old tasks to the new one without catastrophic forgetting.

In general, continual learning methods can be summarized into the following three categories:
(1) \textit{regularization}-based methods~\cite{kirkpatrick2017overcoming, lee2017overcoming, chaudhry2018riemannian,dhar2019learning, ritter2018online, schwarz2018progress, zenke2017continual} that aim at  preserving the most important parameters for previous tasks under a fixed model capacity, by constraining the weight update through model regularization; (2)
\textit{memory}-based methods~\cite{shin2017continual, wu2018memory,riemer2018learning} that preserve the performance of previous tasks by replaying their data during learning of new tasks, in the format of either real data or synthetic data from generative models; (3) \textit{structure}-based methods~\cite{rusu2016progressive,li2017learning,fernando2017pathnet,rosenfeld2018incremental,hung2019compacting,yoon2017lifelong, li2019learn, veniat2020efficient} that dynamically expand the network capacity to reduce the interference between the new tasks and the previously learned ones. Besides, some other methods have been proposed (e.g., \cite{rebuffi2017icarl,pomponi2020efficient}) based on the combination of the above strategies.
In this work, we focus on the structure-based methods due to the following reasons: (1) The regularization approaches tend to forget the learned skills gradually, because the training data of previous tasks is not available when learning new tasks, and the network capacity is fixed (thus limited). 
(2) The memory-based approaches have a general issue that they require explicit re-training using old information accumulated, which could lead to either large working memory or compromise between the information memorized and forgetting.


\begin{figure*}
    \centering
    \includegraphics[width=0.75\linewidth]{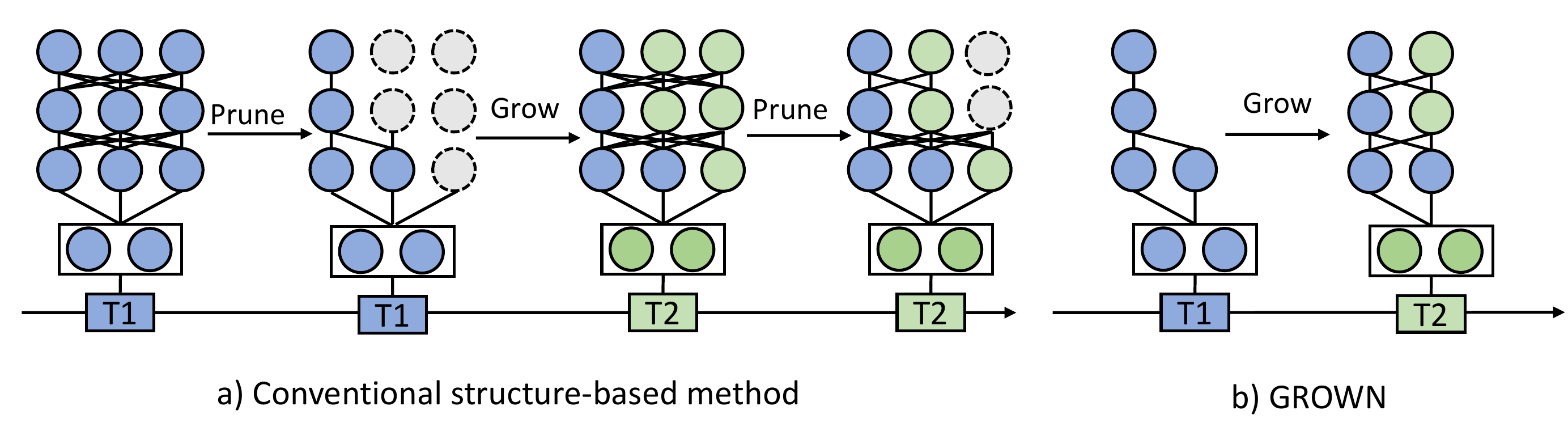}
    \vspace{-1em}
    \caption{The workflow of the proposed GRWON method against with conventional structure-based method \cite{yoon2017lifelong,hung2019compacting}. a) The conventional structure-based method starts from a full-sized model, and then needs to gradually prune-and-grow for each task; b) Our proposed GROWN method starts from a small seed model and grows only when necessary for each task. }
\label{fig:methods_comp}
\end{figure*}

In general, by expanding and pruning the full-sized backbone network, current structure-based methods can achieve better performance than the regularization- and memory-based methods, which however may incur significant computation cost to learn the ideal structure.
For example, Dynamic-Expansion Net (DEN)\cite{yoon2017lifelong} and Compacting-Picking-Growing
(CPG) \cite{hung2019compacting} combine the model pruning, weight selection and model expansion methods, which gradually prune the task-shared weights and then learn additional task-specific weights. 
\cite{veniat2020efficient, li2019learn} adapt additional searching methods to search the optimal model structure for each task, which is also time-consuming. 
The reason why these methods require the additional pruning or searching step is that they start from a full-sized network which is redundant in nature.
Thus inspired, we advocate to 
\emph{learn the best structure for a task in an opposite way by growing from a small seed network, instead of training and reducing from a large full-sized network. }

Intuitively, growing from a small seed network could significantly reduce the computation cost by eliminating the unnecessary pruning used in training a full-sized network. Nevertheless, without careful design, the network may grow aggressively when learning new tasks and quickly reach the full-sized backbone, eventually evolving as in traditional structure-based approaches. In fact, how to appropriately expand the network for learning new tasks, i.e., when to grow, how to grow, and when to stop, has been a long-standing open problem in structure-based CL approaches, where only several heuristic approaches are proposed. For instance, \cite{hung2019compacting}  adds filters or nodes in the network layers and then prunes to make the model compact. The redundant growth not only introduces additional computation cost but also interferes the learning performance, calling for a more elegant growing method to achieve efficient CL. 


To address the above challenges, we develop 
a learnable sparse growth method for efficient model expansion, building on which a novel end-to-end structure-based continual learning framework, \emph{GROWN}, is devised to \emph{dynamically grow the model only when necessary}. More specifically, different from \emph{all} previous structure-based CL methods that prune a full-sized network, our GROWN method starts from a basic seed network with a much smaller size as shown in \cref{fig:methods_comp}. When a new task arrives, the model weights for the old tasks are fixed similar as in previous structure-based methods. Next, we pick and reuse part of the current network: (1) some of the old-task weights that are critical to the new task via a differentiable mask, and (2) the weights that are released during sparse model growth of previous tasks. If the accuracy goal is not attained yet, indicating growing is needed, the proposed learnable sparse growth method is utilized to expand the current model and resume the procedure. 
As the model weights for old tasks are only picked and fixed, we can integrate the required function mappings in a compact model without affecting their accuracy in inference, therefore addressing the catastrophic forgetting issue. By growing from a small network and expanding the model only when necessary, the computation cost is significantly reduced,  while the learning performance is guaranteed by introducing more degree of freedom for optimization via the learnable sparse growth method.
In summary, our main contributions of this work include:
\begin{itemize} 
    \item As an attempt to address the problem of how to appropriately expand the network in CL, we propose a novel learnable sparse growth method to eliminate the unnecessary model growth, which is amenable to integration with general structure-based approaches. 
    \item Building on the proposed learnable sparse growth method, we next develop a novel end-to-end learning framework for efficient continual learning, namely GROWN, which \emph{starts from a small seed network} and \emph{grows only when necessary}.
    \item We conduct extensive experiments to corroborate the superiority of GROWN over current structure-based approaches in multiple CL benchmarks, and characterize the relationship between growth ratio and accuracy for each tasks.
\end{itemize}




\section{Related Work}
\subsection{Continual learning}
\subsubsection{Regularization-based method.}
To address catastrophic forgetting, regularization-based methods protect the old tasks by adding regularization terms in the loss function that constrain the change of neural network weights. Multiple approaches have been proposed, such as Elastic Weight Consolidation (EWC)~\cite{kirkpatrick2017overcoming}, Synaptic Intelligence (SI)~\cite{zenke2017continual}, and Memory Aware Synapses (MAS)~\cite{aljundi2018memory}. The main idea behind these approaches is to estimate the importance of each weight with respect to the trained task. During the training of a new task, any change to the important weights of the old tasks is penalized. 
Despite that regularization methods are suitable for the situation where one can not access the data from previous tasks, their performance degrade quickly in the classical lifelong learning scenario.

\subsubsection{Memory-based method.}
Memory-based methods replay the old tasks data along with the current task data to mitigate the catastrophic forgetting of the old tasks. Deep Generative Replay (DGR)~\cite{shin2017continual} trains a generative model on the data distribution instead of storing the original data from previous tasks. Similar work has been done by Mocanu et al.~\cite{ritter2018online}. Other methods combine the rehearsal and regularization strategies, such as iCaRL~\cite{rebuffi2017icarl}, which exploits a distillation loss along with an examplar set to impose output stability of old tasks. The main drawbacks of memory-based methods are the memory overhead of storing old data or a model for generating them, the computational overhead of retraining the data from all previous tasks, and the unavailability of the previous data in some cases.

\subsubsection{Structure-based method.}
Structure-based approaches \cite{rusu2016progressive,li2017learning,fernando2017pathnet,rosenfeld2018incremental,hung2019compacting,yoon2017lifelong, li2019learn, veniat2020efficient} adapt the network architecture with a sequence of tasks. PNN~\cite{rusu2016progressive} expands the architecture for new tasks and keeps the function mappings by preserving the previous weights. LWF \cite{li2017learning} divides the model layers into two parts, i.e., the shared part and the task-specific part, where the former is co-used by tasks and the later grows with further branches for new tasks. DAN\cite{rosenfeld2018incremental} extends the architecture per new task, where each layer in the new-task model is a sparse linear-combination of the original filters in the corresponding layer of a base model. These methods can considerably mitigate or avoid catastrophic forgetting via architecture expansion, but the model is monotonically increased, leading to a redundant structure. In contrast to directly expanding model architecture, \cite{yoon2019scalable} adds additional task-specific parameters for each task and selectively learns the task-shared parameters together. \cite{li2019learn} adapts architecture search to find the optimal structure for each of the sequential tasks. 

As continually growing the architecture retains the model redundancy, some approaches, e.g., Dynamic-expansion Net (DEN) \cite{yoon2017lifelong}, perform model compression before expansion to obtain a more compact model. Particularly, DEN reduces the weights of the previous tasks via sparse-regularization. Newly added weights and old weights are both adapted for the new task with sparse constraints. 
CPG \cite{hung2019compacting} combines the model pruning, weight selection and model expansion methods, which gradually prunes the task-shared weights and then learns additional task-specific weights. \cite{veniat2020efficient} leverages a task-driven prior over the exponential search space of all possible ways to combine modules, enabling efficient learning on long streams of tasks.

\subsection{Model Growth}
Network Morphism \cite{wei2016network} searches for efficient deep networks by extending layers while preserving the parameters. Recently proposed Autogrow \cite{wen2020autogrow} takes an AutoML approach to grow layers. These methods either require a specially-crafted policy to stop growth (e.g., after a fixed number of layers) or rely on evaluating accuracy during training, incurring significant computational cost. To address this problem, \cite{yuan2020growing} proposes to grow efficient deep networks via structured continual sparsification, which decreases the computational cost  of both inference and training. The method is simple to implement and quick to execute; it automates the network structure reallocation process under practical resource budgets. 

\section{Learnable Sparse Growth}

As alluded to earlier, it is critical to appropriately expand the model in structure-based methods for continual learning, and the basic idea behind the current heuristic approaches is to simply add filters or nodes in the network layers, followed by pruning to make the model compact. Such a two-stage approach clearly incurs additional computation cost and may also fail to find the optimal structure for the current task. Inspired by the learnable growth method \cite{yuan2020growing} in neural architecture search, we propose a novel and provenly better approach, namely \emph{learnable sparse growth}, to enable sparse model growth just as what is needed for learning a single task.

\begin{figure*}
    \centering
    \includegraphics[width=0.3\linewidth]{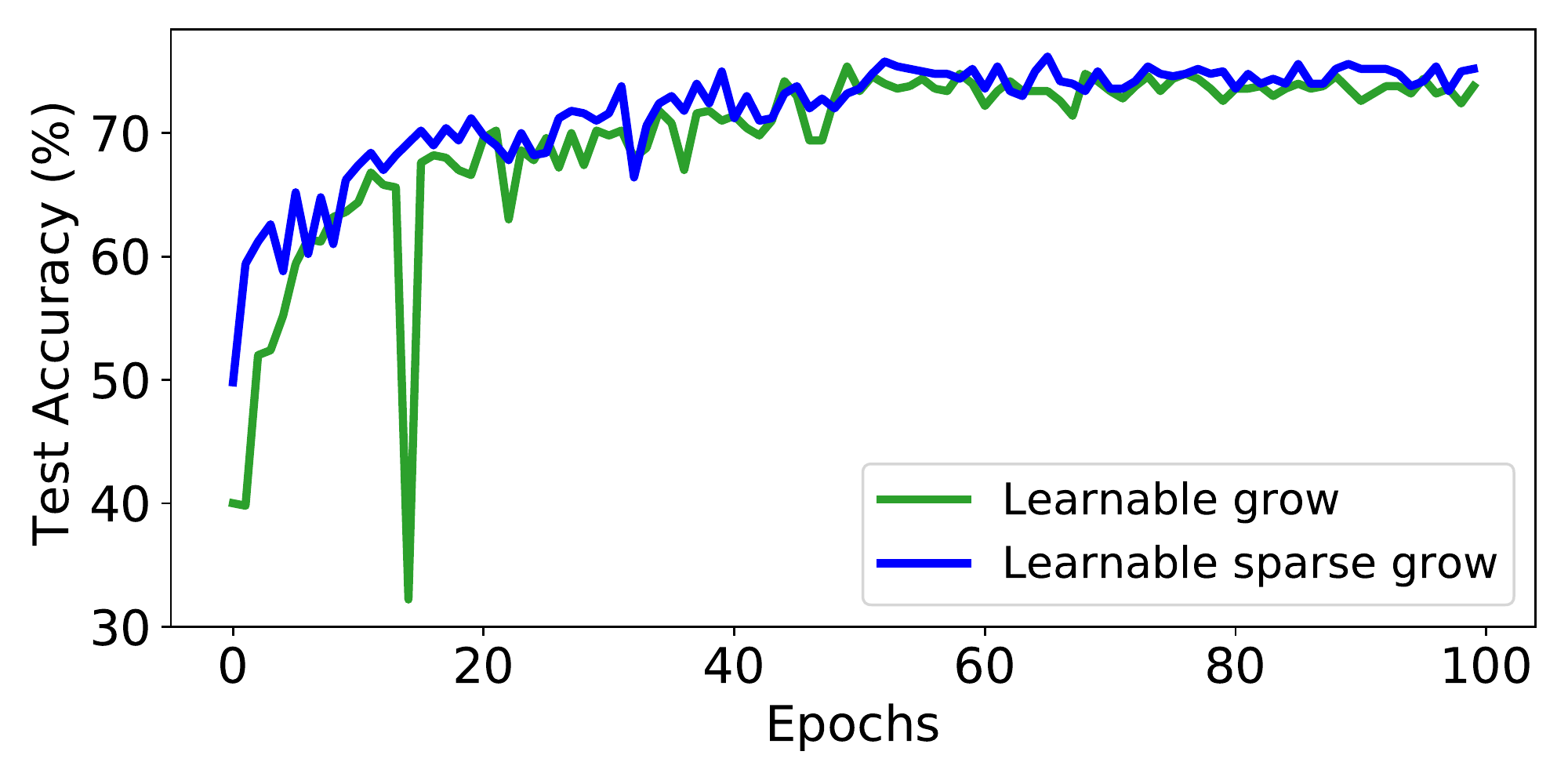} 
    \includegraphics[width=0.3\linewidth]{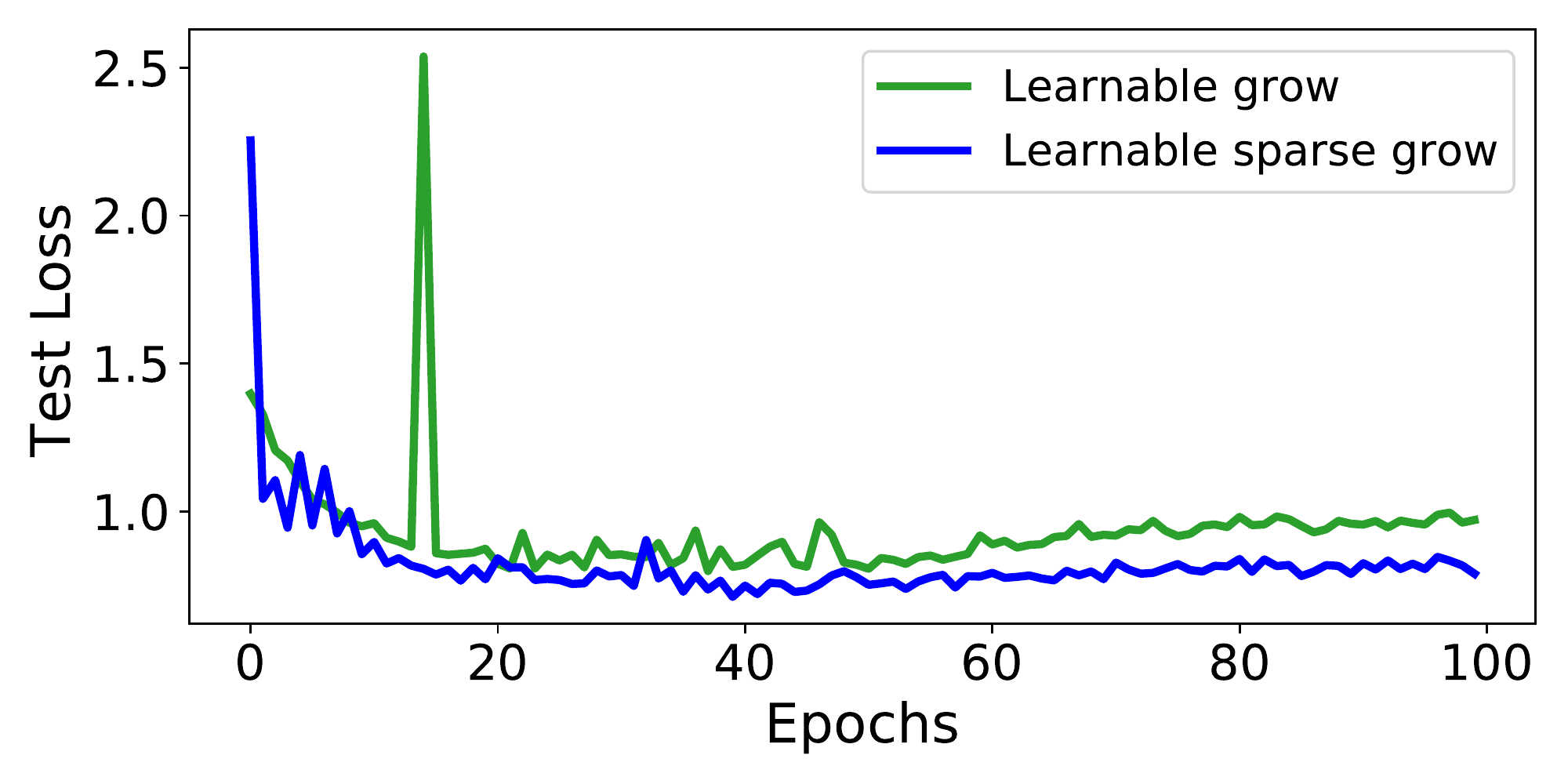} 
    \includegraphics[width=0.3\linewidth]{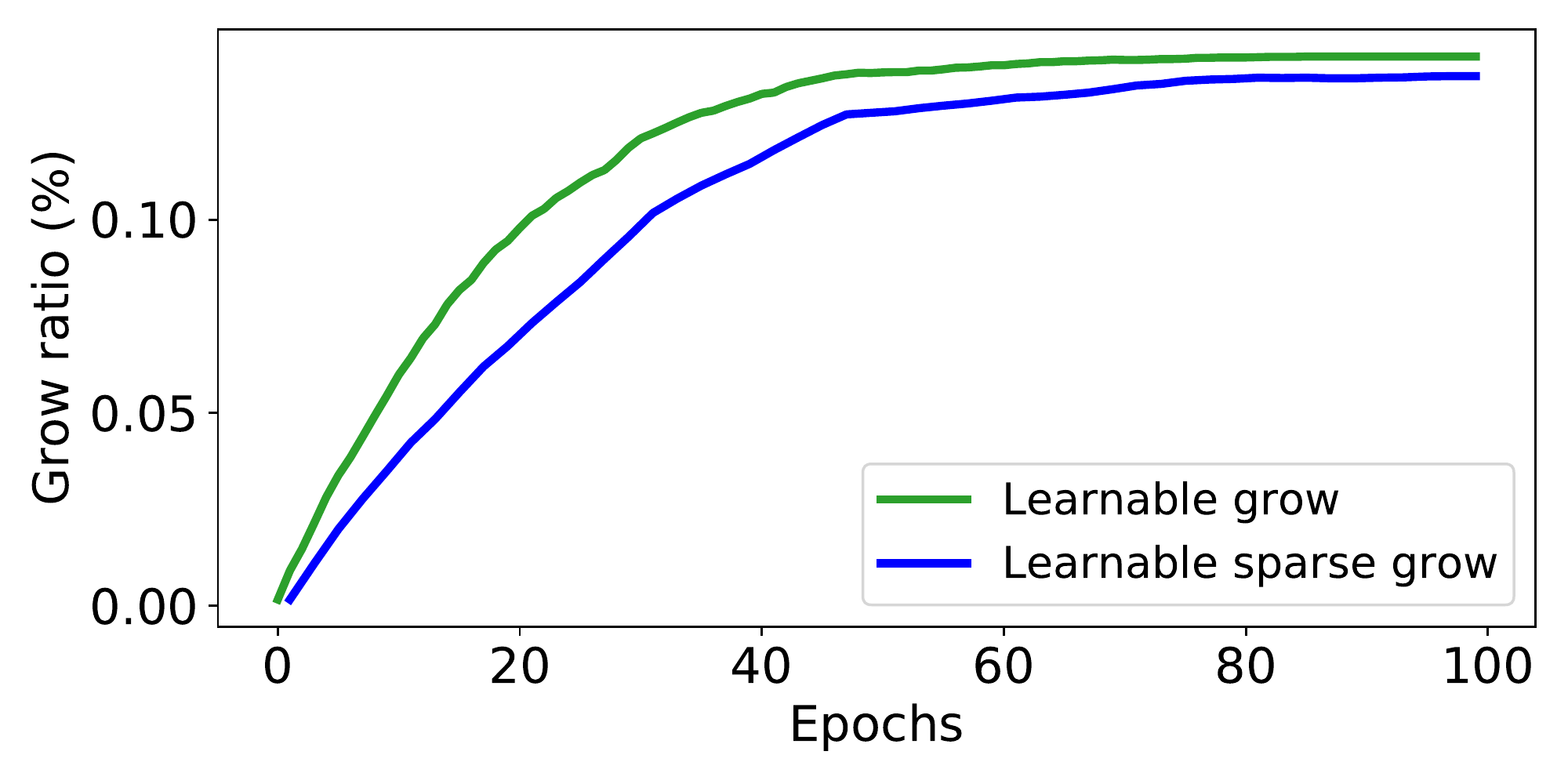}
    \vspace{-1em}
    \caption{Training curve of learnable growth \cite{yuan2020growing} vs our learnable sparse growth on one task of CIFAR-100 Superclass benchmark by using VGG16-BN model. Left: the test accuracy curve; Middle: the test loss curve; Right: the grow ratio curve.}
\label{fig:sparse_growth}
\end{figure*}

\subsection{Revisiting Learnable Growth}
\label{sec:learn_g}
Recently, \cite{yuan2020growing} proposes a method to dynamically grow deep networks by continually re-configuring their architecture during training, which aims to reduce the computational cost for both training and inference. Such a re-configuring process is achieved by learning a channel-wise mask, namely \textbf{grown-mask}, $\vm_g=\{\vm_{g,n}\}_{n=1}^N$ for a $N$-layer convolutional neural network, where each binary element $\vm_{g,n}^i \in \{0, 1\}$ is associated with a channel, to enable training-time pruning ($\vm_{g,n}^i = 1 -> 0$) and growing ($\vm_{g,n}^i = 0 -> 1$) dynamics. Specifically, for the $n$-th convolution layer with $l_{in,n}$ input channels, $l_{out,n}$ output channels and $k \times k$ sized kernels, the $i$-th output feature $x_{out,n}^{i}$ is computed based on the $i$-th filter and the input feature $x_{in,n}$, i.e., for $i \in \{1, ..., l_{out,n}\}$:
{\small
\begin{equation}
    x_{out,n}^{i} = f(x_{in,n}, F_n^{i}\cdot \vm_{g,n}^{i})
\label{eqt:conv_gr}
\end{equation}}%
where $f$ is the operation of the convolutional layers, and $\vm_{g,n}^{i}$ is shared across the filter and broadcast to the same shape as the filter tensor $F_n^{i}$, for enabling pruning/growing the output channel. The growing scheme starts from a slim channel configuration, queries the binary mask parameter and then performs state transitions: (1) When flipping a binary mask element from 0 to 1 for the first time, we grow a randomly initialized filter and concatenate it to the network. (2) If an element flips from 1 to 0, we temporarily detach the corresponding filter from the computational graph. It will be grown back to the  original position if the element flips back to 1; otherwise, it will be permanently pruned at the end of training. (3) For other cases, the corresponding filters either survive and continue training or remain detached pending the next query to the corresponding mask elements. 

The learnable mask variable can be jointly optimized with weights $\vw$ using data $(\vx,\vy)$, which is formulated as follows:
{\small
\begin{equation}
\mathcal{L}(\vw,\vm^g;\vx{}) =
    \mathcal{L}(f(\{\vw \odot \vm^{g} \};\vx), \vy) + \lambda ||\vm^{g}||_{0}
\label{eqt:learn_grow}
\end{equation}}%
where $f$ is the operation in \cref{eqt:conv_gr}. $\vw \odot \vm^{g}$  is a general expression of growing channels and $\mathcal{L}$ denotes a loss function (e.g., cross-entropy loss for classification). The $l_{0}$ term encourages the sparsity of the grown-mask $\vm^g$ so as to limit the grow strength, and $\lambda$ is a coefficient scaling factor. 

\subsection{Learnable Sparse Growth (LSG)}

Traditional learnable growth works well for training a single model on general image classification dataset (e.g., CIFAR-10, ImageNet), which however may fail in continual learning due to the following reasons. (1) \emph{Unstable Training.} We observe that unstable training issue occurs for the learnable growth method on small dataset in the setting of CL. 
\cref{fig:sparse_growth} (Left and Middle) shows the accuracy and loss curve  for one single task of CIFAR-100 Superclass setting, which divides the whole dataset into 20 tasks and each task has only 5 classes, 2500 training images and 500 testing images. It can be easily seen that both the accuracy and the test loss have large perturbations in the earlier epochs, and the test loss increases in the later training epochs. The main reason why learnable growth causes unstable training is that the fine-grained channel-wise growth involves too much of large new initial learnable parameters, which may significantly change the current optimization point. (2) \emph{Overly aggressive growth.} As shown in \cref{fig:sparse_growth} (Right), the learnable growth method has a larger grow ratio for a single task but no better accuracy is achieved.
Such a redundant model growth is clearly not preferable in continual learning, as the backbone network would aggressively expand with the number of learnt tasks increasing.

\begin{figure}[t]
    \centering
    \includegraphics[width=0.75\linewidth]{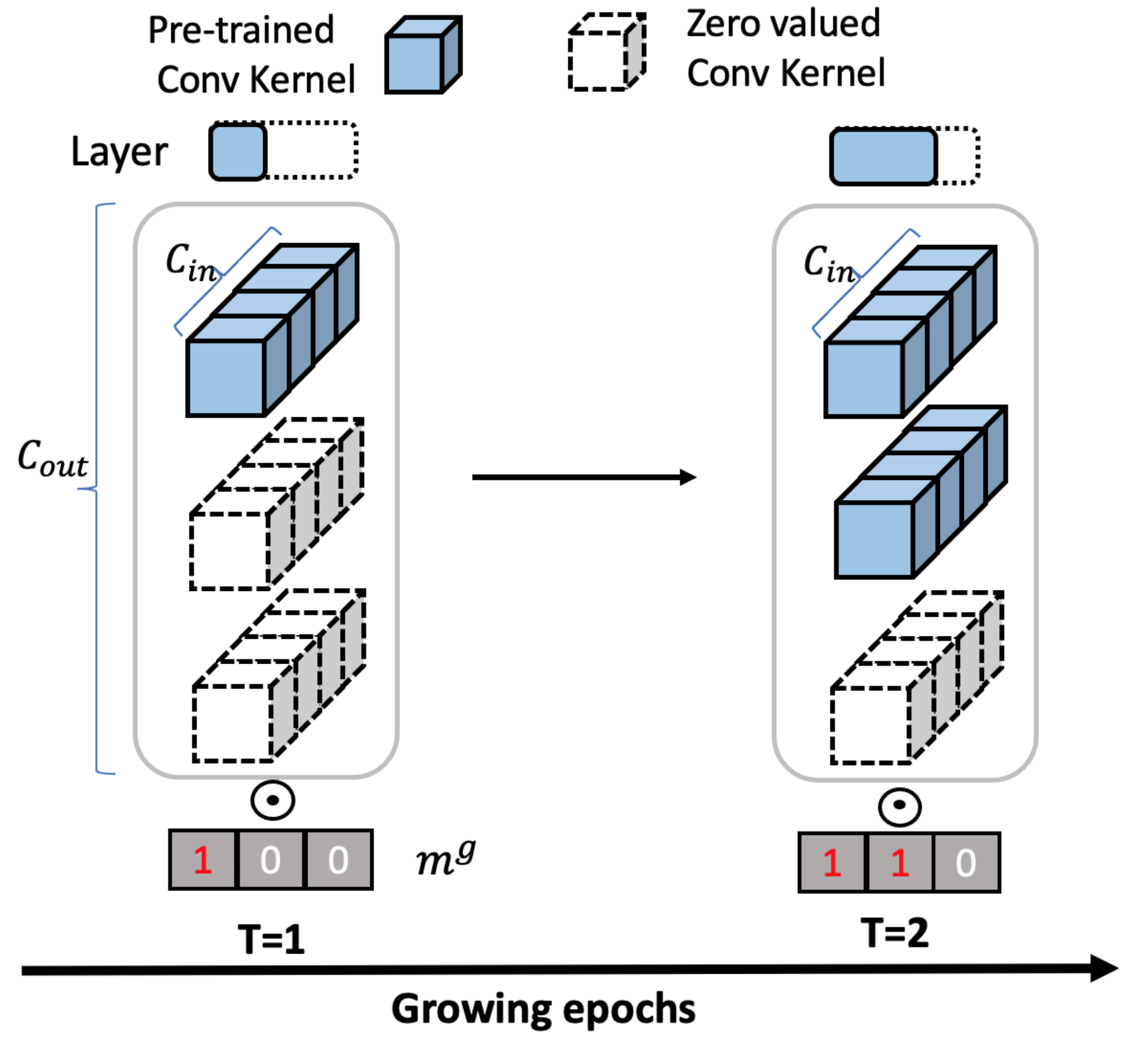} \\
    (a) Conventional learnable growth \\
    \vspace{+1em}
    \includegraphics[width=0.75\linewidth]{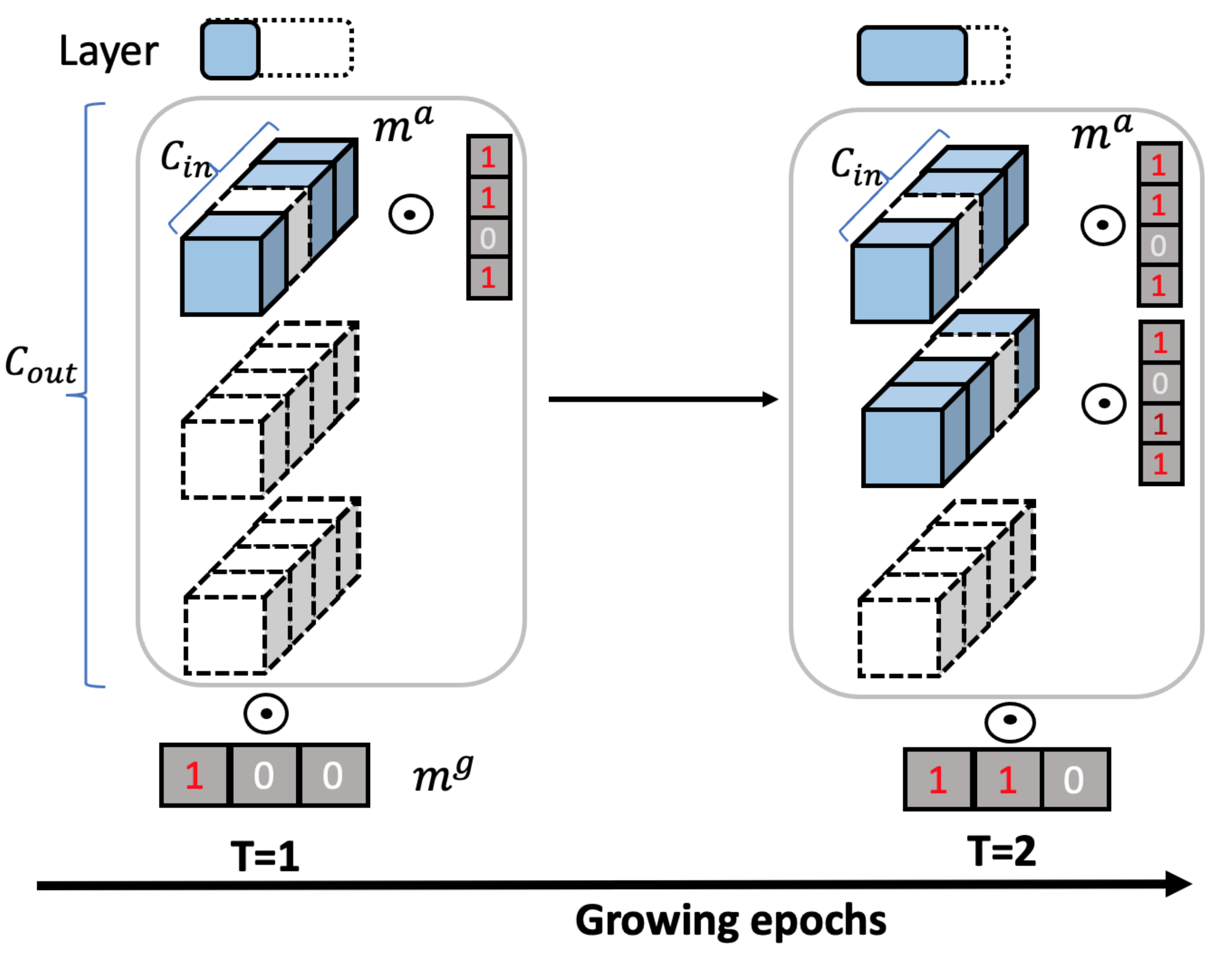} \\
    (b) Learnable sparse growth\\ 
    \caption{Comparison between (a) conventional learnable growth \cite{yuan2020growing} and (b)  Learnable sparse growth.}
\label{fig:sparse_s_growth}
\end{figure}

To address these important issues, we propose a novel \emph{learnable sparse grow} method, by introducing a learnable kernel-wise mask $\vm^{a}$ to selectively pick the kernels among all  grown channels, named as \textbf{attentive-mask}. As shown in \cref{fig:sparse_s_growth}, the kernel-wise mask $\vm^{a}$ keeps the same dimension with the grown-mask $\vm^{g}$,
and introduces sparsity in the grown weights, which can be jointly optimized as follows:
{\small
\begin{equation}
\mathcal{L}(\vw,\vm^g,\vm^a;\vx) =
    \mathcal{L}(f(\{\vw \odot \vm^{g} \odot \vm^{a} \};\vx), \vy) + \lambda ||\vm^{g}||_{0}.
\label{eqt:learn_s_sparse}
\end{equation}}%
Intuitively, minimizing \cref{eqt:learn_s_sparse} explicitly optimizes the model growth in a way that only important and necessary channels for learning the current task will be added to the backbone network. 

To better understand the advantage of introducing the attentive-mask $\vm^a$ for enabling learnable sparse growth, we further analyze the learning performance of \cref{eqt:learn_s_sparse} from the viewpoint of optimization, which is summarized in the following \cref{prop1}.

\begin{proposition}\label{prop1}
Denote $\mathcal{L}(\vw_1^*,(\vm^g)^*_1,(\vm^a)^*_1)=\min_{\vw,\vm^g,\vm^a} \mathcal{L}(\vw,\vm^g,\vm^a;\vx)$ and $\mathcal{L}(\vw_2^*,(\vm^g)^*_2)=\min_{\vw,\vm^g} \mathcal{L}(\vw,\vm^g;\vx)$. It can be shown that
{\small
\begin{align*}
    \mathcal{L}(\vw_1^*,(\vm^g)^*_1,(\vm^a)^*_1)\leq \mathcal{L}(\vw_2^*,(\vm^g)^*_2).
\end{align*}}%
\end{proposition}

\textbf{Remarks.} Intuitively, Proposition~\ref{prop1} indicates that  the proposed learnable sparse growth method can always achieve lower loss values compared to the learnable growth in \cite{yuan2020growing}. This performance improvement can be further interpreted from the following perspectives:
(1) The conventional learnable growth as shown in \cref{eqt:learn_grow} is mathematically equivalent to the following reformulation by multiplying an identity tensor:
{\small
\begin{equation}
\mathcal{L}(\vw,\vm^g;\vx) =
    \mathcal{L}(f(\{\vw \odot \vm^{g} \odot \mathbb{1} \};\vx), \vy) + \lambda ||\vm^{g}||_{0}.
\end{equation}}%
That is to say, \cref{eqt:learn_grow} implicitly puts a constraint on $\vm^a$ that $\vm^a=\mathbb{1}$, whereas in \cref{eqt:learn_s_sparse} the attentive-mask $\vm^a$ can be optimized over a   larger space with more degree of freedom. (2) From the perspective of optimization, without the sparse attentive-mask $\vm^a$, the learnt model weights $\vw_2^*$ cannot be zero for the grown channels induces by the grown-mask $\vm_g$. In stark contrast, the weights in $\vw_1^*$ that correspond to the grown channels can still be zero if they are masked out by the sparse attentive-mask $\vm^a$, leading to more degree of freedom for optimizing the model weights $\vw$.

\section{Continual Learning based on Learnable Sparse Growth}
Clearly, the learnable sparse growth can work independently and serve as a general remedy for expanding the network model in a systematic way. Taking advantage  of the learnable sparse growth, we next propose an end-to-end structure-based method, \emph{GROWN}, for continual learning by growing from a small seed network only when necessary. 
Particularly, we present our method in the sequential-task manner.

\textbf{Learning Task 1:~}
Similar to training a network for a single dataset, given the first task, we start from a basic seed network and then gradually grow it by using the proposed learnable sparse grown method as shown in \cref{eqt:learn_s_sparse}. After training, the current model will serve as the backbone model for next task.

\begin{figure}[t]
    \centering
    \includegraphics[width=1.0\linewidth]{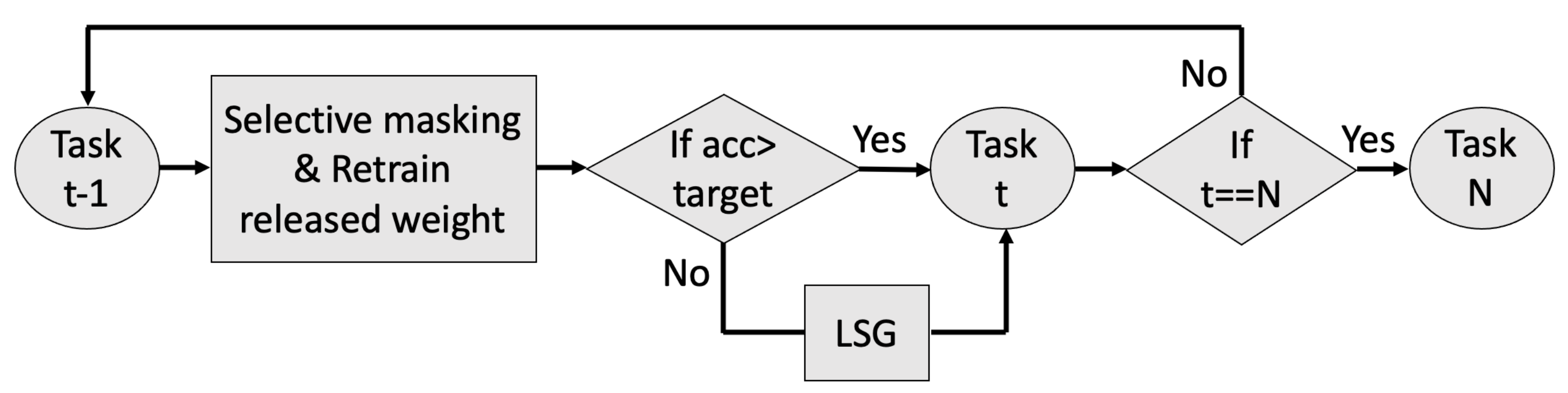}
    \caption{Flowchart of the continual learning framework based on learnable sparse growth. }
\label{fig:overview}
\end{figure}

\textbf{Learning Tasks 2, ..., T:~} 
Assume that in task $t$, the model that can handle task 1 to $t-1$ has been built. We follow a two-stage strategy to learn the model for task $t$: 
\begin{enumerate}
    \item \textbf{Pick and reuse.} Before growing the model, we first utilize two techniques to adapt the preserved model for previous tasks to current task: 1) \textbf{Selective masking}: inspired by recent mask-based multi-task adaption method~\cite{yang2021ksm}, we apply a learnable soft kernel-wise mask to the preserved model, so as to select the important weights for current task; 2) \textbf{Re-train the released weights}: benefiting from the learnable sparse growth method, the preserved model contains sparse weights that can be re-trained for current task without interfering previous tasks. Note that these two techniques are independent and hence can be jointly optimized in the same training process.
    \item  \textbf{Expand.} After training, if the current accuracy is lower than the target accuracy, we will adapt the proposed learnable sparse growth method to integrate more but only necessary features to the current model.
\end{enumerate}
The overview of the framework is shown in \cref{fig:overview}. In what follows, we will present each technique in detail.

\subsection{Selective masking on previous tasks}
To mitigate catastrophic forgetting and utilize the useful information from previous tasks, we first freeze the model weights, and only learn a binary mask to select the important weights for current task. Inspired by \cite{yang2021ksm}, we apply a soft kernel-wise mask, named as \textbf{selective-mask} $\vm^s$ to maximise the utilization of the preserved weights, which can be formulated as:
{\small
\begin{equation}
\mathcal{L}_{t}(\vw_t;\vx{_t}) =
    \mathcal{L}_{t}(f(\{\vw_{t-1} \odot \vm_{t}^{s} \};\vx_{t}), \vy_{t}).
\label{eqt:sel_mask}
\end{equation}}%

The conventional way~\cite{mallya2018piggyback} of generating the binary trainable mask is to train a learnable real-valued mask ($\vm^{r}$) followed by a hard threshold function (e.g., sign function) to binarize it.
Different from that, in this work, we adopt a method to better estimate the gradient by using Gumbel-Sigmoid trick. 

First, we relax the hard threshold function to a continual sigmoid function $\sigma(\vm^{r})$.
Then, to learn the binary mask, we leverage the Gumbel-Sigmoid trick, inspired by Gumbel-Softmax~\cite{jang2017categorical} that performs a differential sampling to approximate a categorical random variable. Since sigmoid can be viewed as a special two-class case of softmax, we define $p(\cdot)$ using the Gumbel-Sigmoid trick as:
{\small
\begin{equation}
    p(\vm^{r}) = \frac{ \textrm{exp}((\textrm{log} \pi_0+g_0)/T)}{\textrm{exp}((\textrm{log}\pi_0+g_0)/T) + \textrm{exp}((g_1)/T)},
\label{eqt:soft_trick}
\end{equation}}%
where $\pi_{0}$ represents $\sigma(\vm^{r})$, $g_0$ and $g_1$ are sampled from Gumbel distribution. The temperature $T$ is a hyper-parameter to adjust the range of input values, 
where choosing a larger value could avoid gradient vanishing during back-propagation. Note that the output of $p(\vm^{r})$ becomes closer to a Bernoulli sample as $T$ is closer to 0. To represent $p(\vm^{r})$ as binary format $\vm^{b}$, we use a hard threshold (i.e., 0.5) during forward-propagation of training. 


\subsection{Re-train the released weights}

As a byproduct of the proposed learnable sparse growth method, the preserved model contains unused weights (i.e., mask as zero) that could be further re-trained for the current task without any impact on previous tasks. Specifically, let $\vw_{1:t-2}$ denote the preserved weights for task $t-2$ and $\vw_{t-2:t-1}$ denote the grown weights for task $t-1$. As we adapt a new kernel-wise attentive-mask $\vm^a$ for the grown weights, $\vw_{t-2:t-1} \times \vm^a$ is utilized by task $t-1$, while the rest can be released for learning the current task $t$.

In addition, it is worthy to note that these two techniques, selective masking and re-train the released weights, work independently on different parts of the weights, and hence can be jointly optimized in a single training process. The loss function can be formulated as:
{\small
\begin{equation}
\mathcal{L}_{t}(\vw_t;\vx{_t}) =
    \mathcal{L}_{t}(f(\{\vw_{t-1} \odot \vm_{t}^{a}, \vw_{t}  \};\vx_{t}), \vy_{t}).
\label{eqt:re_train}
\end{equation}}%

\begin{table*}[!htbp]
\centering
\caption{Accuracy on CIFAR-100 Split 10 tasks}
\label{tab:cifar_spli_acc}
\scalebox{0.75
}{
\begin{tabular}{|l|l|l|l|l|l|l|l|l|l|l|l|l|}
\toprule
Methods & 1 & 2 & 3 & 4 & 5 & 6 & 7 & 8 & 9 & 10 & Avg & Model Size \\ \midrule
Scratch & 62.4 & 57.0 & 67.8 & 64.4 & 67.8 & 67.2 & 68.6 & 67.6 & 61.4 & 69.6 & 63.75  & 10x \\
\midrule \midrule
PNN & 58.2 & 45.6 & 59.4 & 46.8 & 53.2 & 58.0 & 61.2 & 60.4 & 53.8 & 62.4 & 55.90 &  1.7x \\ \midrule
Grow & 56.4 & 45.4 & 56.6 & 47.8 & 53.2 & 55.6 & 61.8 & 57.4 & 53.2 & 59.8 & 54.72 & 1.3x\\ \midrule
DEN & \textbf{61.3} & 49.8 & 60.2 & 49.4 & 54.5 & 57.1 & 64.3 & 62.7 & 55.1 & 66.5 & 58.09 & 1.8x \\ \midrule
CPG & 59.4 & \textbf{55.2} & 64.8 & 57.4 & 59.6 & 59.4 & \textbf{66.2} & 62.8 & 57.6 & 65.4 & 60.12 & 1.3x\\ \midrule
APD & 55.1 & 52.5 & 62.3 & 56.7 & \textbf{61.4} & 58.2 & 63.4 & \textbf{63.1} & 54.8 & 64.2 & 57.54 & 1.3x\\ \midrule
Ours & 57.6 & 53.4 & \textbf{65.2} & \textbf{58.8} & 60.4 & \textbf{60.6} & 65.4 & 62.2  & \textbf{58.0} & \textbf{66.4} & \textbf{60.83} &  \textbf{1.2x} \\ \bottomrule
\end{tabular}}
\end{table*}

\begin{table*}[!htbp]
\centering
\caption{Model size on CIFAR-100 Split 10 tasks}
\label{tab:cifar_spli_size}
\scalebox{0.75
}{
\begin{tabular}{|l|l|l|l|l|l|l|l|l|l|l|}
\toprule
Methods & 1 & 2 & 3 & 4 & 5 & 6 & 7 & 8 & 9 & 10  \\ \midrule
Scratch & 1x & 2x & 3x & 4x & 5x & 6x & 7x & 8x & 9x & 10x  \\
\midrule \midrule
PNN & 1x & 1.08x & 1.16x & 1.24x & 1.32x & 1.40x & 1.48x & 1.54x & 1.62x & 1.7x   \\ \midrule
Grow & 0.5x & 0.64x & 0.78x & 0.86x & 0.92x & 0.99x &  1.06x & 1.16x & 1.24 & 1.3x  \\ \midrule
DEN & 1.0x & 1.2x & 1.46x & 1.46x & 1.64x & 1.64x & 1.8x & 1.8x & 1.8x & 1.8x \\ \midrule
CPG & 1.0x & 1.3x & 1.3x & 1.3x & 1.3x & 1.3x & 1.3x & 1.3x & 1.3x & 1.3x \\ \midrule
APD & 1.0x & 1.05x & 1.10x & 1.13x & 1.16x & 1.19x & 1.23x & 1.26x & 1.28x & 1.3x \\ \midrule
Ours & \textbf{0.5x} & \textbf{0.64x} & \textbf{0.78x} & \textbf{0.78x} & \textbf{0.84x} & \textbf{0.90x} & \textbf{0.90x} & \textbf{1.10x} & \textbf{1.15x} & \textbf{1.2x} \\ \bottomrule
\end{tabular}}
\end{table*}

\subsection{Learnable sparse growth}

As shown in \cref{fig:overview}, after selective masking and re-training the released weights, if the current accuracy is lower than the target accuracy, we will grow the current model by using the proposed learnable sparse growth method to expand the task-specific features. Meanwhile, for the preserved weights, we also adopt the selective masking and re-train the released weights at the same time. The final optimization objective can be formulated as: 
{\small
\begin{equation}
\begin{gathered}
\mathcal{L}_{t}(\vw_t;\vx{_t}) = 
    \mathcal{L}_{t}(f(\{\vw_{t-1} \odot \vm_{t}^{s}, \vw_{t} \odot \vm_{t}^{g} \odot \vm_{t}^{a} \};\vx_{t}), \vy_{t})  \\
    + \lambda ||\vm^{g}||_{0}.
\end{gathered}
\label{eqt:final_loss}
\end{equation}}%

\begin{table*}[!htbp]
\centering
\caption{Accuracy on CIFAR-100 Superclass 20 tasks}
\label{tab:cifar_sup_acc}
\scalebox{0.68}{
\begin{tabular}{|l|l|l|l|l|l|l|l|l|l|l|l|l|l|l|l|l|l|l|l|l|l|l|}
\toprule
Methods & 1 & 2 & 3 & 4 & 5 & 6 & 7 & 8 & 9 & 10 & 11 & 12 & 13 & 14 & 15 & 16 & 17 & 18 & 19 & 20 & Avg & Model Size \\ \midrule
Scratch & 65.8 & \textbf{78.4} & 76.6 & 82.4 & 82.2 & 84.6 & 78.6 & 84.8 & 83.4 & 89.4 & 87.8 & 80.2 & 84.4 & 80.2 & 52.0 & 69.4 & 66.4 & 70.0 & 87.2 & 91.2 & 78.8  & 20x \\ \midrule
Fine-tune & 65.2 & 76.1 & 76.1 & 77.8 & 85.4 & 82.5 & 79.4 & 82.4 & 82.0 & 87.4 & 87.4 & 81.5 & 84.6 & 80.8 & 52.0 & 72.1 & 68.1 & 71.9 & 88.1 & 91.5 & 78.6 & 20x\\ \midrule \midrule
Grow & 67.0 & 73.8 & 74.4 & 75.2 & 81.4 & 81.2 & 78.8 & 80.4 & 80.6  & 85.4 & 85.8 & 80.4 & 81.2 & 80.6 & 50.8 & 68.8 & 66.4 & 68.2 & 84.2 & 88.4 & 76.5  & 1.5x \\ \midrule
DEN & 66.4 & 78.0 & 77.4 & 78.8 & 81.6 & 81.8 & 76.0 & 80.4 & 79.8 & 85.0 & 85.2 & 78.8 & 83.2 & 81.6 & 50.4 & 71.2 & 66.8 & 79.4 & 85.0 & 90.2 & 77.4 & 2.1x \\ \midrule
CPG & 67.0 & \textbf{79.6} & {77.2} & \textbf{82.0} & 86.8 & \textbf{87.2} & 82.4 & \textbf{85.6} & 86.4 & \textbf{89.6} & 90.0 & 84.0 & 87.2 & {84.8} & 55.4 & \textbf{73.8} & \textbf{72.0} & {71.6} & \textbf{89.6} & {92.8} & 81.2 & 1.5x \\ \midrule
Ours & \textbf{67.2} & {76.8} & \textbf{79.6} & 81.8 & \textbf{87.4} & {86.8} & \textbf{84.2} & 83.8 & \textbf{87.8} & {89.4} & \textbf{91.0} & \textbf{84.6} & \textbf{87.2} & \textbf{85.0} & \textbf{55.4} & 73.6 & {71.2} & \textbf{73.8} & {89.2} & \textbf{94.6} & \textbf{82.2}  &  1.5x\\ \bottomrule
\end{tabular}}
\end{table*}

\begin{table*}[!htbp]
\centering
\caption{Model size on CIFAR-100 Superclass 20 tasks}
\label{tab:cifar_sup_size}
\scalebox{0.63}{
\begin{tabular}{|l|l|l|l|l|l|l|l|l|l|l|l|l|l|l|l|l|l|l|l|l|l|}
\toprule
Methods & 1 & 2 & 3 & 4 & 5 & 6 & 7 & 8 & 9 & 10 & 11 & 12 & 13 & 14 & 15 & 16 & 17 & 18 & 19 & 20  & Model Size \\ \midrule
Scratch & 1x & 2x & 3x & 4x & 5x & 6x & 7x & 8x & 9x & 10x & 11x & 12x & 13x & 14x & 15x & 16x & 17x & 18x & 19x & 20x & 20x\\ \midrule
Fine-tune & 1x & 2x & 3x & 4x & 5x & 6x & 7x & 8x & 9x & 10x & 11x & 12x & 13x & 14x & 15x & 16x & 17x & 18x & 19x & 20x  & 20x\\ \midrule \midrule
Grow &  0.3x & 0.4x & 0.48x & 0.56x & 0.65x & 0.75x & 0.75x & 0.83x & 0.94x & 0.94x & 1.12x & 1.18x &  1.24x & 1.24x & 1.30x & 1.30x & 1.38x & 1.44x & 1.44x & 1.5x & 1.5x \\ \midrule
DEN & 1.0x & 1.06x & 1.13x & 1.13x & 1.19x & 1.25x & 1.32x & 1.32x & 1.40x & 1.46x & 1.46x & 1.55x & 1.61x & 1.61x & 1.61x & 1.70x & 1.82x & 1.90x & 2.02x & 2.1x &   2.1x \\ \midrule
CPG & 1.0x & 1.5x & 1.5x & 1.5x & 1.5x & 1.5x & 1.5x & 1.5x & 1.5x & 1.5x & 1.5x&  1.5x & 1.5x & 1.5x & 1.5x & 1.5x  & 1.5x & 1.5x & 1.5x & 1.5x &  1.5x \\ \midrule
Ours &  \textbf{0.3x} & \textbf{0.4x} & \textbf{0.48x} & \textbf{0.56x} & \textbf{0.65x} & \textbf{0.75x} & \textbf{0.75x} & \textbf{0.83x} & \textbf{0.94x} & \textbf{0.94x} & \textbf{1.12x} & \textbf{1.18x} &  \textbf{1.24x} & \textbf{1.24x} & \textbf{1.30x} & \textbf{1.30x} & \textbf{1.38x} & \textbf{1.44x} & \textbf{1.44x} & 1.5x & 1.5x\\ \bottomrule
\end{tabular}}
\end{table*}

\begin{table*}[!htbp]
\centering
\caption{Ablation study on CIFAR-100 Superclass 20 tasks}
\label{tab:ablation}
\scalebox{0.72}{
\begin{tabular}{|l|l|l|l|l|l|l|l|l|l|l|l|l|l|l|l|l|l|l|l|l|l|}
\toprule
Methods & 1 & 2 & 3 & 4 & 5 & 6 & 7 & 8 & 9 & 10 & 11 & 12 & 13 & 14 & 15 & 16 & 17 & 18 & 19 & 20 & Avg  \\ \midrule
Grow only & 67.0 & 73.8 & 74.4 & 75.2 & 81.4 & 81.2 & 78.8 & 80.4 & 80.6  & 85.4 & 85.8 & 80.4 & 81.2 & 80.6 & 50.8 & 68.8 & 66.4 & 68.2 & 84.2 & 88.4 & 76.5  \\ 
+Sel-mask & +0.0 & +2.0 & +3.4 & +4.0 & +3.4 & +4.2 &  +3.6 & +2.0 & +3.8 & +3.0 & +3.8 & +3.0 & +3.6 & +3.4 & +2.8 & +4.0 & +3.2 & +3.8 & +3.0 & +3.6 & +3.7  \\
+LSG & +0.2 & +0.6 & +1.0 & +1.8 & +1.4 & +0.8 & +1.2 & +0.6 & +2.2 & +0.6 & +1.0 & +0.6 & +2.0 &+0.4 & +1.4 & +0.6 & +0.6 & +1.0 & +1.4 & +1.6 & +1.3  \\ 
+Re-train & +0 & +0.4 & +0.8 & +0.4 & +0.4 & +0.6 & +0.6 & +0.8 & +1.0 & +0.4 & +0.8 & +0.6 & +0.4 & +1.2 & +0.4 & +0.2 & +1.0 & +0.8 & +0.6 & +1.0 & +0.7  \\ \midrule
Ours & {67.2} & {76.8} & 79.6 & 81.8 & {86.8} & {86.8} & {84.2} & 83.8 & {87.8} & {89.4} & {91.0} & {84.6} & {87.2} & 85.0 & {55.4} & 73.6 & {71.2} & {73.8} & {89.2} & 94.6 & {82.2} \\ \bottomrule
\end{tabular}}
\end{table*}

\section{Experiments}
\subsection{Experimental settings}
We evaluate our GROWN on multiple datasets against state-of-the-art continual learning methods.

\textbf{1) CIFAR-100 Split}. CIFAR-100 \cite{krizhevsky2009learning} consists of images from 100 generic object classes. We split the classes into 10 group, and consider 10-way multi-class classification in each group as a single task. We use 5 random training/validation/test splits of 4000/1000/1000 samples. We use a modified version of LeNet-5 with 20-50-800-500 neurons as the base model and train 20 epochs for each task sequentially. 

\textbf{2) CIFAR-100 Superclass}. We divide the CIFAR-100 dataset into 20 tasks. Each task has 5 classes, 2500 training images, and 500 testing images. In the experiment, VGG16-BN model (VGG16 with batch normalization layers) is employed to train the 20 tasks sequentially. 

\textbf{3) MiniImageNet Split}. We divide the MiniImageNet dataset into 20 tasks. Each task has 5 classes, 2375 training images and 500 testing images. In the experiment, ResNet18 is employed to train the 20 tasks sequentially. Due to the space limitation, we relegate the experimental results on MiniImageNet to the appendix.

\subsection{Methods for comparison}
To test the efficacy of our method, we compare it with several representative methods in three categories:

\textbf{1) Baselines}: We adapt two regular training schemes as follows and select the best accuracy as the target for our method: (1) scratch that we train the model from scratch for each task individually (Scratch); fine-tune that we train the model from scratch for the first
task only and then fine-tune it for the rest tasks. 

\textbf{2) Grow only}: This method only grows the model for each task without effect on the pre-trained backbone part during training. We compare with two related works, PNN \cite{li2017learning} which linearly grows the model for each task, and Grow~\cite{yuan2020growing} which grows the model by learnable mask as mentioned earlier.

\textbf{3) Grow-and-prune}: This method gradually grows and prunes the backbone model for each new task. We choose two representative works for comparison, DEN \cite{yoon2017lifelong}, CPG \cite{hung2019compacting}. In addition, we also compare APD \cite{yoon2019scalable}, which grows the parameter size by involving the task-specific parameters with L1 norm constrain. 

\subsection{Quantitative evaluation}
\subsubsection{Accuracy comparison.} 
We first validate the performance on CIFAR-100 Split dataset as shown in \cref{tab:cifar_spli_acc}. 
The accuracy of training from scratch for each task invidiously serves as the target accuracy for our method. Compared with Grow Only methods - PNN and Grow, our method could significantly improve the accuracy for each task, with an even smaller model size. In addition, compared with Grow-and-prune methods - DEN and CPG which require one order more training time, our method achieves better results than DEN on all tasks and CPG on most of tasks. Similar phenomenon can be observed with CIFAR-100 Superclass results as shown in \cref{tab:cifar_sup_acc}. Compared with the best previous results CPG, we achieve 1.0$\%$ accuracy gain in average.

\subsubsection{Model size comparison.}
For a fair comparison, all the methods are trained on the single NVIDIA Quadro RTX 5000 GPU with the same batch size (i.e, 32). \cref{tab:cifar_spli_size} and \cref{tab:cifar_sup_size} summarizes the model size for each task during training on CIFAR-100 Split and Superclass respectively. It's worthy to note that different from all compared works, our method could start from a compact model instead of a full size model. By doing so, our method could guarantee the smallest model size not only for the final task, but also for each task during training.





\section{Ablation Study and Analysis}

\subsection{The effect of each component in GROWN}
We study the effectiveness of each components for the proposed learnable sparse growth in continual learning on CIFAR-100 Superclass setting. As shown in \cref{tab:ablation}, we consider four different combinations to perform this ablation study: 1) Only grow the model for each task by learnable mask~\cite{yuan2020growing}, without any updating on the pre-trained part (Grow only); 2) Adapt the selective masking techniques on top of the grow only (Sel-mask); 3) Replace the grow only with our proposed learnable sparse growth (LSG); 4) Further add the re-train the released weights technique (Re-train) to show the final version. First, it  can be seen that the selective-masking method largely improves the accuracy for each task, which shows that simply masking unimportant weights could achieve high adaption ability. Second, our proposed learnable sparse growth (LSG) method can not only improve the accuracy, but also reduce the model size by involving kernel-wise sparsity. Last, benefiting from LSG which masks the room to re-train the released weights, it could further improve 0.7$\%$ accuracy in average for all tasks.

\section{Conclusion}
In this work, we propose GRWON, a novel end-to-end continual learning framework to \textit{dynamically grow the model only when necessary}. Different from all previous structure-based methods, the GROWN starts from a small seed network instead of a full-sized one. Furthermore, to efficiently and appropriately expand network structure for new task, we develop a learnable sparse growth method eliminating the additional pruning/searching step in previous structure-based method. We validate GROWN on multiple datasets against
state-of-the-art methods, which shows superiority improvement in both accuracy and model size. In particular, we achieve 1.0$\%$ accuracy gain in average compared to the current SOTA results on CIFAR-100 Superclass 20 tasks setting. 

\clearpage
\small{
\bibliography{reference.bib}
}

\clearpage
\appendix


\end{document}


\section{Appendix}
\subsection{Algorithm}

\begin{algorithm}[!htbp]
\SetAlgoLined
\kwInput{Given a pre-defined model with initialized weights $\vw$, grown-mask $\vm^{g}$, kernel-wise attentive-mask $\vm^{a}$, selective-mask $\vm^{s}$}
\eIf{task t == 1}{
  Set a target accuracy for task 1 \\
  Apply the attentive-mask $\vm_{1}^{a}$ and grown-mask $\vm_{1}^{g}$ to sparse grow the current model as Eq.(3)\;
}{
\For{task t = 2 ... T}{
    Set a target accuracy for task t \\
    Apply the selective-mask $\vm_{t}^{s}$ for fixed $\vw_{t-1}^{f}$, and retrain the released weights $\vw_{t-1}^{r}$ as Eq.(5) \\
    \eIf{Current accuracy $<$ Target accuracy}{
    Apply the selective-mask $\vm_{t}^{s}$ and grown-mask $\vm_{t}^{g}$ to sparse grow the model $\vw_{t-1}$ to obtain $\vw_{t}$ with growing ratio constraint as Eq.(8)}{
    Set current model as $\vw_{t}$}
    }
}
\caption{GROWN}
\label{alg:sparse_g}
\end{algorithm}

\subsection{Proof of Proposition 1}

The conventional learnable growth as shown in \cref{eqt:learn_grow} is equal to multiply an identity tensor:
\begin{align*}
&\mathcal{L}(\vw, \vm^g, \vm^a=\mathbb{1};\vx)\\
=& \mathcal{L}_{t}(f(\{\vw \odot \vm^{g} \odot \mathbb{1} \};\vx), \vy) + \lambda ||\vm^{g}||_{0}.
\end{align*}
For ease of exposition, let $\Tilde{\vw}=(\vw,\vm^g)$. Denote
\begin{align*}
    (\Tilde{\vw}^*_1,(\vm^a)_1^*)&=\arg\min \mathcal{L}(\Tilde{\vw},\vm^a);\\
    \Tilde{\vw}^*_2&=\arg\min \mathcal{L}(\Tilde{\vw},\vm^a=\mathbb{1}).
\end{align*}
Then, it follows that
\begin{align*}
    &\mathcal{L}(\Tilde{\vw}^*_1,(\vm^a)_1^*)\\
    \leq& \min\{\mathcal{L}(\Tilde{\vw},\vm^a|\vm^a=\mathbb{1}), \mathcal{L}(\Tilde{\vw},\vm^a|\vm^a\neq\mathbb{1})\}
\end{align*}
for any $\Tilde{\vw},\vm^a$.

For $\Tilde{\vw}=\Tilde{\vw}^*_2$, 
\begin{itemize}
    \item if there does not exist any $\vm^a\neq \mathbb{1}$, such that $\mathcal{L}(\Tilde{\vw}^*_2,\vm^a|\vm^a\neq\mathbb{1}) \leq \mathcal{L}(\Tilde{\vw}^*_2,\vm^a|\vm^a=\mathbb{1})$, then
    \begin{align*}
    &\mathcal{L}(\Tilde{\vw}^*_1,(\vm^a)_1^*)\\
    \leq& \min\{\mathcal{L}(\Tilde{\vw},\vm^a|\vm^a=\mathbb{1}), \mathcal{L}(\Tilde{\vw},\vm^a|\vm^a\neq\mathbb{1})\}\\
    \leq& \min\{\mathcal{L}(\Tilde{\vw}^*_2,\vm^a|\vm^a=\mathbb{1}), \mathcal{L}(\Tilde{\vw}^*_2,\vm^a|\vm^a\neq\mathbb{1})\}\\
    =&\mathcal{L}(\Tilde{\vw}^*_2,\vm^a=\mathbb{1});
\end{align*}
    
    \item if there exists some $\vm^a_2\neq \mathbb{1}$, such that $\mathcal{L}(\Tilde{\vw}^*_2,\vm^a_2) \leq \mathcal{L}(\Tilde{\vw}^*_2,\vm^a=\mathbb{1})$, then
    \begin{align*}
    &\mathcal{L}(\Tilde{\vw}^*_1,(\vm^a)_1^*)\\
    \leq& \min\{\mathcal{L}(\Tilde{\vw},\vm^a|\vm^a=\mathbb{1}), \mathcal{L}(\Tilde{\vw},\vm^a|\vm^a\neq\mathbb{1})\}\\
    \leq& \min\{\mathcal{L}(\Tilde{\vw}^*_2,\vm^a=\mathbb{1}), \mathcal{L}(\Tilde{\vw}^*_2,\vm^a_2\}\\
    =&\mathcal{L}(\Tilde{\vw}^*_2,\vm^a_2)\\
    \leq& \mathcal{L}(\Tilde{\vw}^*_2,\vm^a=\mathbb{1}).
\end{align*}
\end{itemize}
Therefore, we can conclude that $\mathcal{L}(\Tilde{\vw}^*_1,(\vm^a)_1^*)\leq \mathcal{L}(\Tilde{\vw}^*_2,\vm^a=\mathbb{1})$.

\subsection{Results on miniImageNet dataset}

\begin{table*}[!htbp]
\centering
\caption{Accuracy on miniImageNet split 20 tasks}
\scalebox{0.7}{
\begin{tabular}{|l|l|l|l|l|l|l|l|l|l|l|l|l|l|l|l|l|l|l|l|l|l|l|}
\toprule
Methods & 1 & 2 & 3 & 4 & 5 & 6 & 7 & 8 & 9 & 10 & 11 & 12 & 13 & 14 & 15 & 16 & 17 & 18 & 19 & 20 & Avg & Model Size \\ \midrule
Scratch & 66.0 & 69.2 & 65.4 & 63.2 & 70.6 & 73.0 & 67.0 & 64.6 & 65.8 & 64.4 & 66.2 & 68.2 & 63.2 & 70.0 & 66.2 & 69.6 & 70.2 & 66.8 & 68.6 & 65.2 & 67.5  & 20x \\ \midrule \midrule
Grow & 63.8 & 61.8 & 60.0 & 57.4 & 63.6 & 65.2 & 57.0 & 56.4 & 58.4 & 56.0 & 57.4 & 60.6 & 53.8 & 65.4 & 61.8 & 62.8 & 58.4 & 59.4 & 64.0 & 55.8 & 60.1 & 1.5x \\ \midrule
DEN & 65.4 & 65.0 & 64.6 & 60.0 & 64.8 & 68.6 & 62.4 & 59.8 & 59.4 & 60.2 & 59.4 & 64.6 & 55.8 & 63.8 & 63.4 & 60.4 & 65.2 & 59.6 & 65.8 & 59.2 & 62.5 &  1.9x \\ \midrule
CPG & 64.6 & \textbf{68.6} & 64.2 & 60.4 & 65.0 & 69.2 & \textbf{63.8} & 62.6 & 62.2 & \textbf{63.4} & 66.0 & 65.2 & 59.8 & \textbf{67.0} & 63.2 & 65.8 & 61.6 & 64.0 & \textbf{67.4} & 60.4 & \textbf{69.3} & 1.5x \\ \midrule
Ours & \textbf{65.8} & 68.2 & \textbf{65.4} & \textbf{61.6} & \textbf{68.8} & \textbf{71.2} & \textbf{63.8} & \textbf{63.6} & \textbf{64.2} & 62.8 & \textbf{66.4} & \textbf{66.2} & \textbf{60.6} & 66.0 & \textbf{65.2} & \textbf{66.8} & \textbf{63.2} & \textbf{65.6} & {65.8} & \textbf{63.4} &  {65.4} &  1.5x\\ \bottomrule
\end{tabular}}
\end{table*}

\begin{table*}[!htbp]
\centering
\caption{Model size on miniImageNet split 20 tasks}
\scalebox{0.65}{
\begin{tabular}{|l|l|l|l|l|l|l|l|l|l|l|l|l|l|l|l|l|l|l|l|l|l|}
\toprule
Methods & 1 & 2 & 3 & 4 & 5 & 6 & 7 & 8 & 9 & 10 & 11 & 12 & 13 & 14 & 15 & 16 & 17 & 18 & 19 & 20  & Model Size \\ \midrule
Scratch & 1x & 2x & 3x & 4x & 5x & 6x & 7x & 8x & 9x & 10x & 11x & 12x & 13x & 14x & 15x & 16x & 17x & 18x & 19x & 20x  & 20x \\ \midrule \midrule
Grow only & 0.35x & 0.44x & 0.52x & 0.60x & 0.65x & 0.65x & 0.73x & 0.80x & 0.88x & 0.95x & 1.14x & 1.14x &  1.22x & 1.28x & 1.28x & 1.34x & 1.40x & 1.40x & 1.45x & 1.5x & 1.5x  \\ \midrule
DEN & 1.0x & 1.05x & 1.11x & 1.16x & 1.16x & 1.23x & 1.39x & 1.39x & 1.45x & 1.45x & 1.50x & 1.55x & 1.59x & 1.59x & 1.64x & 1.64x & 1.70x & 1.76x & 1.82 & 1.9x & 1.9x\\ \midrule
CPG & 1.0x & 1.5x & 1.5x & 1.5x & 1.5x & 1.5x & 1.5x & 1.5x & 1.5x & 1.5x & 1.5x&  1.5x & 1.5x & 1.5x & 1.5x & 1.5x  & 1.5x & 1.5x & 1.5x & 1.5x &  1.5x  \\ \midrule
Ours & \textbf{0.35x} & \textbf{0.44x} & \textbf{0.52x} & \textbf{0.60x} & \textbf{0.65x} & \textbf{0.65x} & \textbf{0.73x} & \textbf{0.80x} & \textbf{0.88x} & \textbf{0.95x} & \textbf{1.14x} & \textbf{1.14x} &  \textbf{1.22x} & \textbf{1.28x} & \textbf{1.28x} & \textbf{1.34x} & \textbf{1.40x} & \textbf{1.40x} & \textbf{1.45x} & \textbf{1.5x} & 1.5x \\ \bottomrule
\end{tabular}}
\end{table*}